\documentclass[journal,10pt]{elsarticle}

\usepackage{algorithmic}
\usepackage{array}

\usepackage{framed,multirow,booktabs}
\usepackage{graphicx,epsfig}
\usepackage{amssymb,amsmath,bm}
\usepackage{textcomp}
\usepackage{amsthm}
\usepackage{lineno,hyperref}
\usepackage{xcolor}
\usepackage{cleveref}
\usepackage[section]{placeins}

\journal{Journal of Pattern Recognition}

\begin{document}
\begin{frontmatter}
%
\title{Stroke Constrained Attention Network for Online Handwritten Mathematical Expression Recognition}

\author{Jiaming Wang, Jun Du and Jianshu Zhang}

\begin{abstract}
  In this paper, we propose a novel stroke constrained attention network (SCAN) which treats stroke as the basic unit for encoder-decoder based online handwritten mathematical expression recognition (HMER). Unlike previous methods which use trace points or image pixels as basic units, SCAN makes full use of stroke-level information for better alignment and representation. The proposed SCAN can be adopted in both single-modal (online or offline) and multi-modal HMER. For single-modal HMER, SCAN first employs a CNN-GRU encoder to extract point-level features from input traces in online mode and employs a CNN encoder to extract pixel-level features from input images in offline mode, then use stroke constrained information to convert them into online and offline stroke-level features. Using stroke-level features can explicitly group points or pixels belonging to the same stroke, therefore reduces the difficulty of symbol segmentation and recognition via the decoder with attention mechanism.
  For multi-modal HMER, other than fusing multi-modal information in decoder,
  SCAN can also fuse multi-modal information in encoder by utilizing the stroke based alignments between online and offline modalities. The encoder fusion is a better way for combining multi-modal information as it implements the information interaction one step before the decoder fusion so that the advantages of multiple modalities can be exploited earlier and more adequately when training the encoder-decoder model.
  Evaluated on a benchmark published by CROHME competition, the proposed SCAN achieves the state-of-the-art performance.

\end{abstract}


\end{frontmatter}
{\bf Correspondence:}
Dr. Jun Du, National Engineering Laboratory for Speech and Language Information Processing (NEL-SLIP), University of Science and Technology of China, No. 96, JinZhai Road, Hefei, Anhui P. R. China (Email: jundu@ustc.edu.cn).


\newpage

\section{Introduction}
\label{sec:Introduction}
Handwritten mathematical expression recognition (HMER) is one of the primary branches of document analysis and recognition, which is widely used for the electronization of various scientific literatures. Different from online/offline character or text line recognition, HMER is much more challenging as it meets with the complicated two-dimensional structural analysis~\cite{HmerChallenge1,HmerChallenge2,HmerChallenge3}.

Generally, HMER consists of two major problems~\cite{chan2000mathematical}, which are symbol recognition and structural analysis. Traditional methods solve these problems sequentially or globally. Concretely, sequential methods~\cite{zanibbi2002sequential1,alvaro2014sequential2} first segment input expression into mathematical symbols and identify them separately. Then the structural analysis finds out the structure of the expression according to the symbol recognition results. While global methods~\cite{awal2014global1,alvaro2016global2} deal with HMER as a global optimization of symbol recognition and structural analysis. The symbol segmentation is performed implicitly .

As deep learning came to prominence, attention based encoder-decoder approaches are extensively adopted for HMER, which can be divided into online and offline cases. For online HMER, ~\cite{Zhang2017gruHmer,TAP19} treat the handwritten mathematical expression (HME) as a point sequence and extract point-level features from input traces. While for offline HMER,~\cite{WAP17,Deng2017WYGIWYS} take the HME as a static image and extract pixel-level features from the input image. Benefiting from rich dynamic (spatial and temporal) information which is extremely helpful for handwritten recognition, online HMER tends to meet fewer difficulties caused by ambiguous handwriting. However, the lack of global information in online HMER may lead to incorrect recognition coming from delayed strokes or inserted strokes~\cite{TAP19,WAP17}. On the contrary, offline HMER can easily handle these situations as its input is a static image which contains global information robust to stroke orders. Consequently, it is intuitive to utilize both dynamic traces and static images to build a more powerful recognition system, which is referred as multi-modal HMER~\cite{Wang2019MAN}.

Although encoder-decoder approaches have greatly improved the performance of HMER, they still suffer from the difficulty of symbol segmentation. Because an attention mechanism is utilized to implement symbol segmentation implicitly, the inaccurate attention will lead to the mis-recognition of the input expression. Previous approaches always compute the attention coefficients on low-level features, such as trace points for online modality and image pixels for offline modality. However, for handwriting recognition problems, handwritten input has a distinctive property that points or pixels can be naturally grouped into higher-level basic units, called strokes, formed by a pen-down and pen-up action. Therefore, fully utilizing strokes as the basic units for attention based encoder-decoder models will potentially improve the attention based alignment and even enhance the representation ability of input features for online HMER.


In this study, we propose a novel stroke constrained attention network (SCAN) for online HMER, which treats stroke as the basic unit for encoder-decoder models. It can be adopted in both single-modal and multi-modal HMER. Compared with previous encoder-decoder based approaches \cite{TAP19,WAP17}, SCAN has three major striking properties: (i) It greatly improves the alignment generated by attention; (ii) The number of strokes is much smaller than the number of points or pixels, which helps accelerate the decoding process; (iii) For multi-modal recognition, SCAN provides oracle alignments between online traces and offline images, which enables to fuse features from different modalities in encoder and significantly improves the performance.


\begin{figure*}[htb]
\centerline{\includegraphics[width=0.98\linewidth]{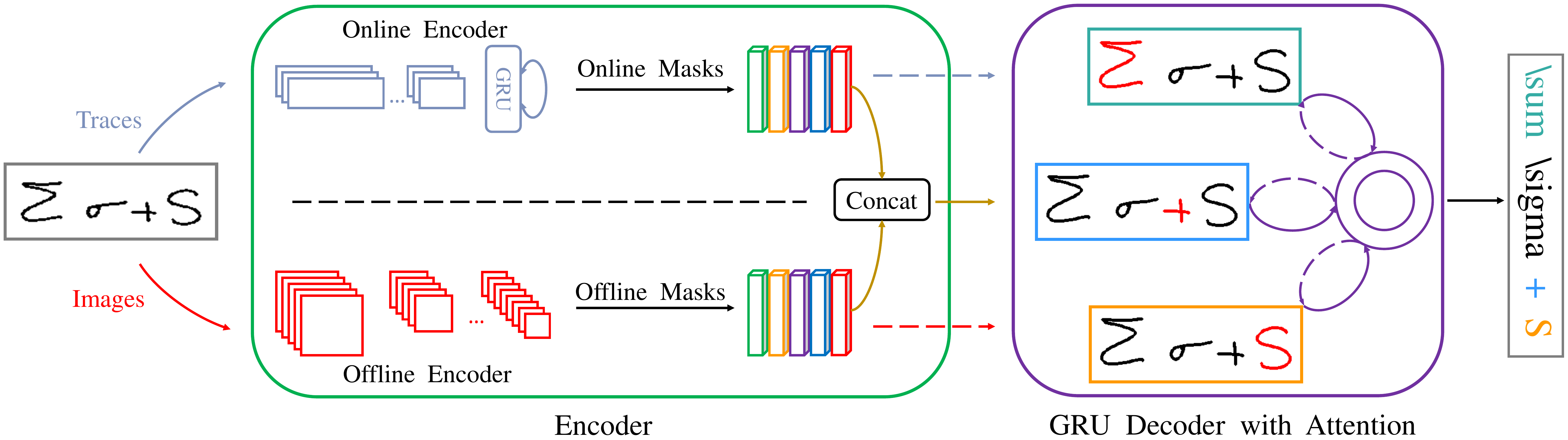}}
\caption{The overall architecture of stroke constrained attention network (SCAN). 
}
\label{overall}
\end{figure*}

As shown in Figure~\ref{overall}, for online modality, we employ a convolutional neural network with gated recurrent units (CNN-GRU) based encoder to extract point-level features from the input trace sequence. Then the stroke constrained information, i.e., the correspondence between points and strokes, is utilized to convert point-level features into online stroke-level features. Similarly, for offline modality, we adopt a CNN-based encoder to extract pixel-level features from the input image and then convert it into offline stroke-level features. A decoder with attention is introduced to generate the recognition result, where attention actually achieves symbol segmentation implicitly. The stroke-level features as a higher-level and more accurate representation extracted from the low-level point/pixel features can potentially reduce the difficulty of symbol segmentation and recognition.

For multi-modal HMER, SCAN can play a more essential role as it not only groups points and pixels into strokes to generate more efficient symbol segmentation, but also makes fusing features from different modalities in encoder become possible. On top of the stroke-level features from both online and offline modalities, we design two multi-modal fusion strategies, namely encoder fusion and decoder fusion. The decoder fusion is similar to our recent work~\cite{Wang2019MAN}, where a multi-modal attention is equipped with re-attention mechanism to guide the decoding procedure by generating a multi-modal stroke-level context vector with the information of both online and offline modalities.
The proposed encoder fusion has one advantage that it takes the information interaction one step before the decoder fusion so that the advantages of multiple modalities can be exploited earlier and more adequately. As~\cite{Wang2019MAN} treats point and pixel as the basic unit, it is difficult to implement the encoder fusion with no explicit alignments between online point-level features and offline pixel-level features. However, as shown in Figure~\ref{overall}, SCAN treats stroke as the basic unit and therefore there are oracle alignments between online and offline stroke-level features. Accordingly, we can fuse them in encoder to obtain multi-modal stroke-level features, which are then fed to the decoder. Finally an attention mechanism is adopted to guide the decoding procedure and generate recognition result step by step. It can utilize both online and offline information to acquire more accurate attention results and significantly improve the performance of online HMER.

The main contributions of this study are summarized as follows:
\begin{itemize}
\item A novel SCAN framework is proposed by fully utilizing the stroke information for encoder-decoder based online HMER.
\item A single-modal SCAN approach is presented via the novel design of online/offline stroke-level features.
\item A multi-modal SCAN approach is introduced with two fusion strategies, namely encoder fusion and decoder fusion.
\item We demonstrate the effectiveness and efficiency of SCAN by attention visualization over the stroke-level features.
\end{itemize}

This work is an extension of our previous conference paper~\cite{Wang2019MAN} in five ways: 1) The stroke as a high-level representation is adopted rather than the point/pixel as a low-level representation in both single-modal and multi-modal HMER; 2) The stroke-level features are used in multi-modal attention equipped with re-attention to show the strength of the stroke constrained information; 3) The online and offline stroke-level features are fully exploited in the encoder fusion strategy; 4) A stroke-level attention guider is proposed to help attention learn better; 5) A comprehensive set of experiments are designed on the published benchmark of CROHME2014/CROHME2016/CROHME2019.

\section{Related Work}
In this section, we first describe traditional approaches and then discuss neural network based approaches for HMER. Finally we elaborate multi-modal machine learning approaches.

\subsection{Traditional approaches for HMER}
One key property of online HMER is that the pen-tip movements (xy-coordinates) and pen states (pen-down and pen-up) can be acquired during the writing process. Traditional approaches for HMER~\cite{stroke_unit1,stroke_unit2,stroke_unit3} usually utilize the pen states to group trajectory points belonging to the same stroke in advance and treat stroke as the basic unit, i.e. representing mathematical expression as a set of strokes. The process of HMER can be divided into two steps: symbol recognition and structural analysis. Symbol recognition involves symbol segmentation and classification. Symbol segmentation is actually grouping the strokes belonging to the same symbol. These two steps can be implemented separately or jointly, referring to sequential and global methods, respectively. Sequential methods~\cite{zanibbi2002sequential1,alvaro2014sequential2} first achieve symbol recognition by finding the best possible groups of strokes and identifying the symbol corresponding to each stroke group. Then structural analysis is performed using syntactic models for representing spatial relations among symbols, such as tree structure models~\cite{pru_a2007tree2}. In sequential methods, the contextual information is not fully exploited and the symbol segmentation/recognition errors will be subsequently propagated to structural analysis. On the contrary, global methods~\cite{awal2014global1,alvaro2016global2} optimize symbol recognition and structural analysis using the complete expression simultaneously. However, global methods are computationally more expensive and efficient search strategies\cite{rhee2009efficient} must be defined.

\subsection{Attention based encoder-decoder approaches for HMER}
Encoder-decoder framework has been extensively applied to many applications including machine translation~\cite{bert,nmt_attention,nmt_attention2}, speech recognition~\cite{asr1,asr2} and image caption~\cite{imageCaption1,imageCaption2,imageCaption3}. Typically, an encoder is first employed to extract high-level representations from input. Then, a decoder is applied to generate a variable-length sequence as the output step by step. To address the issue that both input and output are of variable length, an attention mechanism~\cite{attention1,attention2,attention3} is usually incorporated into decoder, which can generate a fixed-length context vector by weighted averaging the variable-length high-level representations to guide the decoding procedure. With the development of deep learning, encoder-decoder based approaches with an attention mechanism are also widely used for HMER, which convert the output format from tree structure into LaTeX strings and significantly outperform the traditional methods. According to the
different input modalities of HMER, these approaches can be divided into online and offline ones. Online approach treats the HME as dynamic traces while offline approach treats the HME as static images. The online approach~\cite{Zhang2017gruHmer} employed GRU-based encoder and GRU-based decoder with a spatial attention, which achieved significant improvements compared with traditional methods for HMER.~\cite{TAP19} introduced a TAP model with additional temporal attention and an attention guider to further improve the performance. Besides,~\cite{hong2019residual} adopted residual connection in encoder and a transition probability matrix in decoder. As for the offline approach,~\cite{WAP17} utilized a WAP model, which adopted CNN-based encoder to extract features from static images.~\cite{Deng2017WYGIWYS} proposed a coarse-to-fine attention to improve efficiency. In addition,~\cite{wu2018PAL} introduced a PAL model and employed an adversarial learning strategy during training.

\subsection{Multi-modal machine learning}
Recently, an increasing number of studies focus on multi-modal machine learning, which aims to utilize advantages and complementarities from multiple modalities~\cite{multimodal1,multimodal2,multimodal3,multimodal4}. An essential topic of multi-modal is how to fuse the information from different modalities~\cite{multifusion1,multifusion2}. Specific to features with varying length such as sentences, videos and audio streams, one difficulty to make a multi-modal fusion is the unaligned nature of different modalities. As encoder-decoder based framework is widely used for sequence machine learning, here we focus on the discussion of multi-modal fusion in the encoder stage or the decoder stage. Generally, fusing features in encoder can acquire better performance than which in decoder~\cite{DecoderFusionVQA1,DecoderFusionVQA2}, as information from different modalities can interact earlier. However, we usually lack of optimal mapping between different modalities, which makes the encoder fusion challenging. Although \cite{EncoderFusion1,EncoderFusion2} utilized cross-modal self-attention to achieve the encoder fusion, the unaligned issue was still existed as the alignments acquired by cross-modal self-attention could not be guaranteed to be exactly accurate.

Differently, for online HMER, we can obtain oracle alignments between online and offline modalities by making full use of stroke constrained information. Therefore, in this study, we propose SCAN to achieve the fusion of online and offline modalities in encoder, which significantly improves the recognition performance.

\section{Single-modal SCAN}
In this section, we introduce the proposed SCAN for single-modal HMER, including online SCAN (OnSCAN) and offline SCAN (OffSCAN). Different from previous single-modal approaches~\cite{TAP19,WAP17,Deng2017WYGIWYS}, we explicitly utilize the stroke constrained information in encoder-decoder based HMER. Specifically, stroke-level features are adopted in SCAN rather than point-level and pixel-level features. Furthermore, the attention mechanism in the decoder is to discover the alignments between the predicted mathematical symbol and input features. Therefore, the attention in SCAN actually groups strokes belonging to the same symbol, which is obviously much easier and more efficient than grouping points or pixels in previous approaches as stroke-level features are a higher-level representation to reduce the difficulty of attention than the local point-level or pixel-level features.

\subsection{Data Preparation}
For online HMER, the input raw data is the handwritten traces, which can be represented as a variable length sequence:
\begin{equation}
\left \{ \left [ x_{1},y_{1},s_{1} \right ], \left [ x_{2},y_{2},s_{2} \right ],\cdots,  \left [ x_{N},y_{N},s_{N} \right ] \right \}
\end{equation}
where $x_{i}$ and $y_{i}$ are the xy-coordinates of the pen movements and $s_{i}$ indicates which stroke the $i^{\text{th}}$ point belongs to. Please note that in this study the stroke constrained information $\{s_{i}\}$ is always used for both online and offline modalities. The offline modality contains the image which is rendered by lining trace points of each stroke. To convert online point-level or offline pixel-level features to the corresponding stroke-level features, we further generate online and offline stroke masks from the stroke constrained information.

\subsubsection{Input Features}
For the online modality, we normalize the traces and extract an 8-dimensional feature vector for each point $i$:
\begin{equation}
\mathbf{x}_{i}^{\text{on}} = \left [ x_{i},y_{i},\Delta x_{i},\Delta y_{i},\Delta ^{2} x_{i},\Delta ^{2} y_{i},\text{strokeFlag1},\text{strokeFlag2} \right ]
\end{equation}
where $\Delta x_{i}=x_{i+1}-x_{i}$, $\Delta y_{i}=y_{i+1}-y_{i}$, $\Delta ^{2} x_{i}=x_{i+2}-x_{i}$, $\Delta ^{2} y_{i}=y_{i+2}-y_{i}$. The last two terms are flags indicating the status of the pen, i.e., $\left [ 1,0 \right ]$ means pen-down while $\left [ 0,1 \right ]$ means pen-up. We refer to the
trace point sequence after processing as $\mathbf{X}^{\text{on}}=\left \{ \mathbf{x}_{1}^{\text{on}},\mathbf{x}_{2}^{\text{on}},\cdots, \mathbf{x}_{N}^{\text{on}} \right \}$, where $N$ denotes the number of trace points.

For the offline modality, we first calculate the heights of all strokes. Then we compute the average height of strokes with the height greater than one tenth of the maximum height. Furthermore, we normalize xy-coordinates of all points in accordance with the average height and simply line trace points of each stroke to convert traces into static images, $\mathbf{X}^{\text{off}}$ of size ${H}_{\text{in}} \times {W}_{\text{in}}$.

\subsubsection{Stroke Masks}
We believe that the stroke constrained information plays an essential role in online HMER. In~\cite{TAP19,Wang2019MAN}, the stroke information is only used as additional two dimensions of the input 8-dimensional feature vector. So we aim at fully utilizing the stroke information by defining the online and offline stroke masks. Specifically, for one HME sequence, suppose it consists of $M$ strokes and $N$ points. We define online stroke masks as $\mathbf{Mask}^{\text{on}}= \{ \mathbf{mask}^{\text{on}}_{1},\mathbf{mask}^{\text{on}}_{2},\cdots, \mathbf{mask}^{\text{on}}_{M} \}$ and offline stroke masks as $\mathbf{Mask}^{\text{off}}= \{ \mathbf{mask}^{\text{off}}_{1},\mathbf{mask}^{\text{off}}_{2},\cdots, \mathbf{mask}^{\text{off}}_{M} \}$. Each online stroke mask $\mathbf{mask}^{\text{on}}_{j}$ is a $N$-dimension vector and the value of each element $i$ is 1 or 0, indicating whether the $i^\text{th}$ point belongs to the ${j}^\text{th}$ stroke or not by using the original stroke information $\{s_{i}\}$. Each offline stroke mask $\mathbf{mask}^{\text{off}}_{j}$ is a matrix of size ${H}_{\text{in}} \times {W}_{\text{in}}$ and each element $(h,w)$ is 1 or 0, indicating whether the pixel $(h,w)$ belongs to the ${j}^\text{th}$ stroke or not by using the original stroke information $\{s_{i}\}$.

\subsection{Online Encoder}
\label{sec:online encoder}
\begin{figure}[htb]
\centerline{\includegraphics[width=0.95\linewidth]{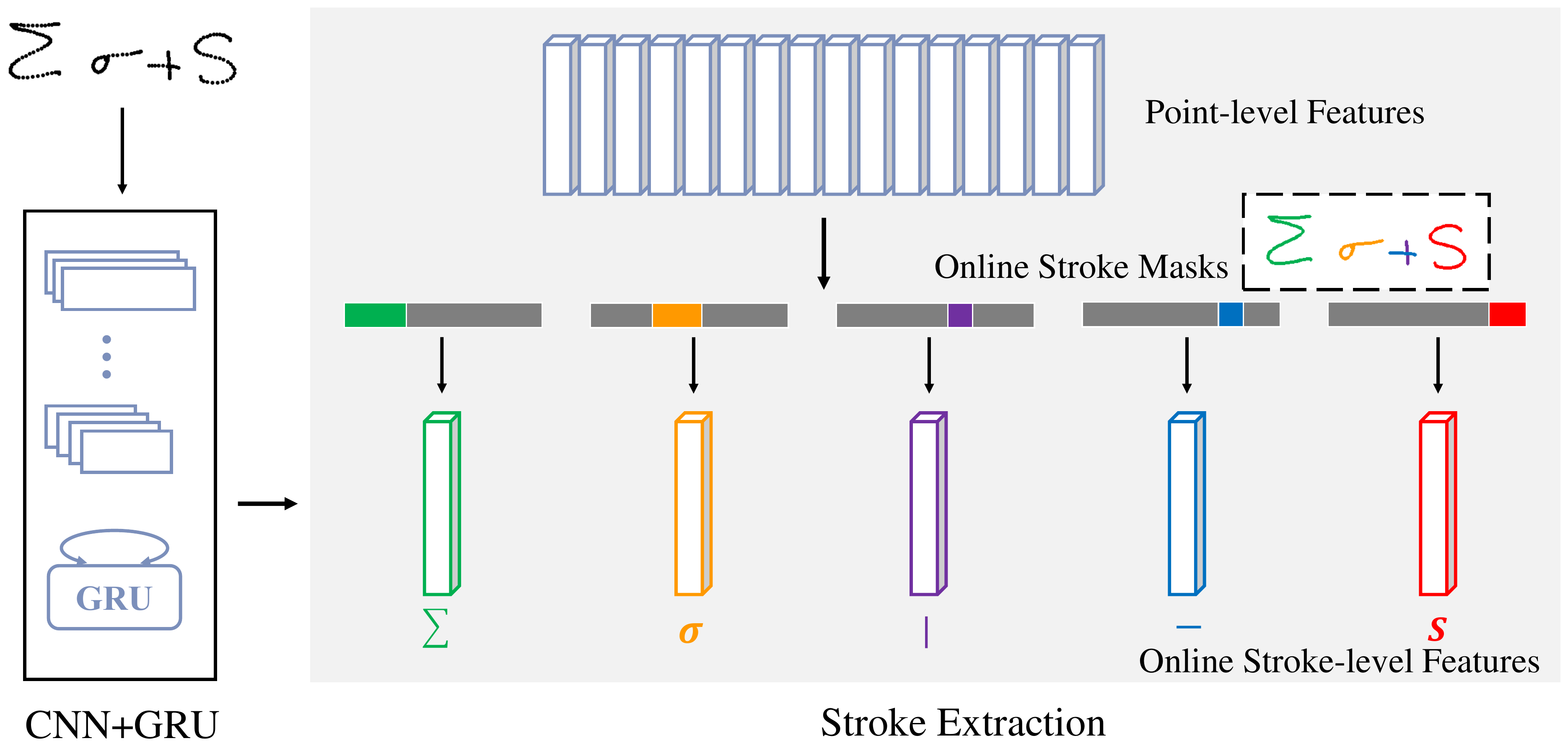}}
\caption{The architecture of online encoder. The left part is point-level feature extraction from input traces using CNN-GRU. The right part is online stroke-level feature extraction from point-level features.}
\label{online-mask}
\end{figure}

The online encoder is designed to extract the online stroke-level features based on $\mathbf{X}^{\text{on}}$ and $\mathbf{Mask}^{\text{on}}$. As shown in the left part of Figure~\ref{online-mask}, different from~\cite{Zhang2017gruHmer,TAP19}, we employ CNN following a fewer stack of GRUs, which can acquire better local information and improve the recognition performance. To match with the CNN input requirement, $\mathbf{X}^{\text{\text{on}}}$ is transformed into a tensor of size $8 \times 1 \times N$, i.e., feature maps with 8 channels and the height set to 1. The convolutional layers of CNN are configured as densely connected layers in DenseNet~\cite{densenet}. Note that we modify the kernel size of convolutional layers from $3 \times 3$ to $1 \times 3$ as the heights of feature maps are 1. The output of CNN is a tensor of size $1 \times L \times D'$, which is then transformed into a $D'$-dimensional vector sequence of length $L$ $\mathbf{A}'=\left \{ \mathbf{a}'_{1},\mathbf{a}'_{2},\cdots, \mathbf{a}'_{L} \right \}$. To capture the context information from input traces, a stack of GRUs are built on top of CNN. The hidden state of GRU can be calculated as:
\begin{equation}
\mathbf{h}'_{t}=\textrm{GRU}\left ( \mathbf{a}'_{t},\mathbf{h}'_{t-1} \right )
\end{equation}
Furthermore, as unidirectional GRU cannot exploit the future context information, we actually adopt bidirectional GRU which can utilize both past and future context information. The detailed implementation of GRU can be found in \cite{TAP19}.

The output of CNN-GRU encoder is a variable-length vector sequence, namely point-level features, which can be represented as $\mathbf{A}=\left \{ \mathbf{a}_{1},\mathbf{a}_{2},\cdots, \mathbf{a}_{L} \right \}$ and each element is a $D$-dimensional vector. Note that $N$ is a multiple of $L$ based on the number of pooling layers in CNN part. With the point-level features, we utilize online stroke masks to convert point-level features into online stroke-level features, which is illustrated in the right part of Figure~\ref{online-mask}. First, the same number of downsampling as that in CNN part of online encoder is used to process online stroke masks, which converts each online mask from a $N$-dimensional vector $\mathbf{mask}^{\text{on}}_{j}$ to a $L$-dimensional vector $\mathbf{pmask}^{\text{on}}_{j}$. Then, the ${j}^{\text{th}}$ online stroke-level feature can be calculated as:
\begin{equation}
\mathbf{s}^{\text{on}}_{j}= \frac{\mathbf{pmask}^{\text{on}}_{j} }{||\mathbf{pmask}^{\text{on}}_{j}||_1} \mathbf{A} \quad \quad \mathbf{S}^{\text{on}}=\left \{ \mathbf{s}^{\text{on}}_{1},\mathbf{s}^{\text{on}}_{2},\cdots, \mathbf{s}^{\text{on}}_{M} \right \}
\end{equation}
where $||\cdot||_1$ is the vector 1-norm, $\mathbf{s}^{\text{on}}_{j}$ is a $D$-dimensional vector and $\mathbf{S}^{\text{on}}$ is the final output of online encoder.

\subsection{Offline Encoder}
\label{sec:offline encoder}
\begin{figure}[htb]
\centerline{\includegraphics[width=0.95\linewidth]{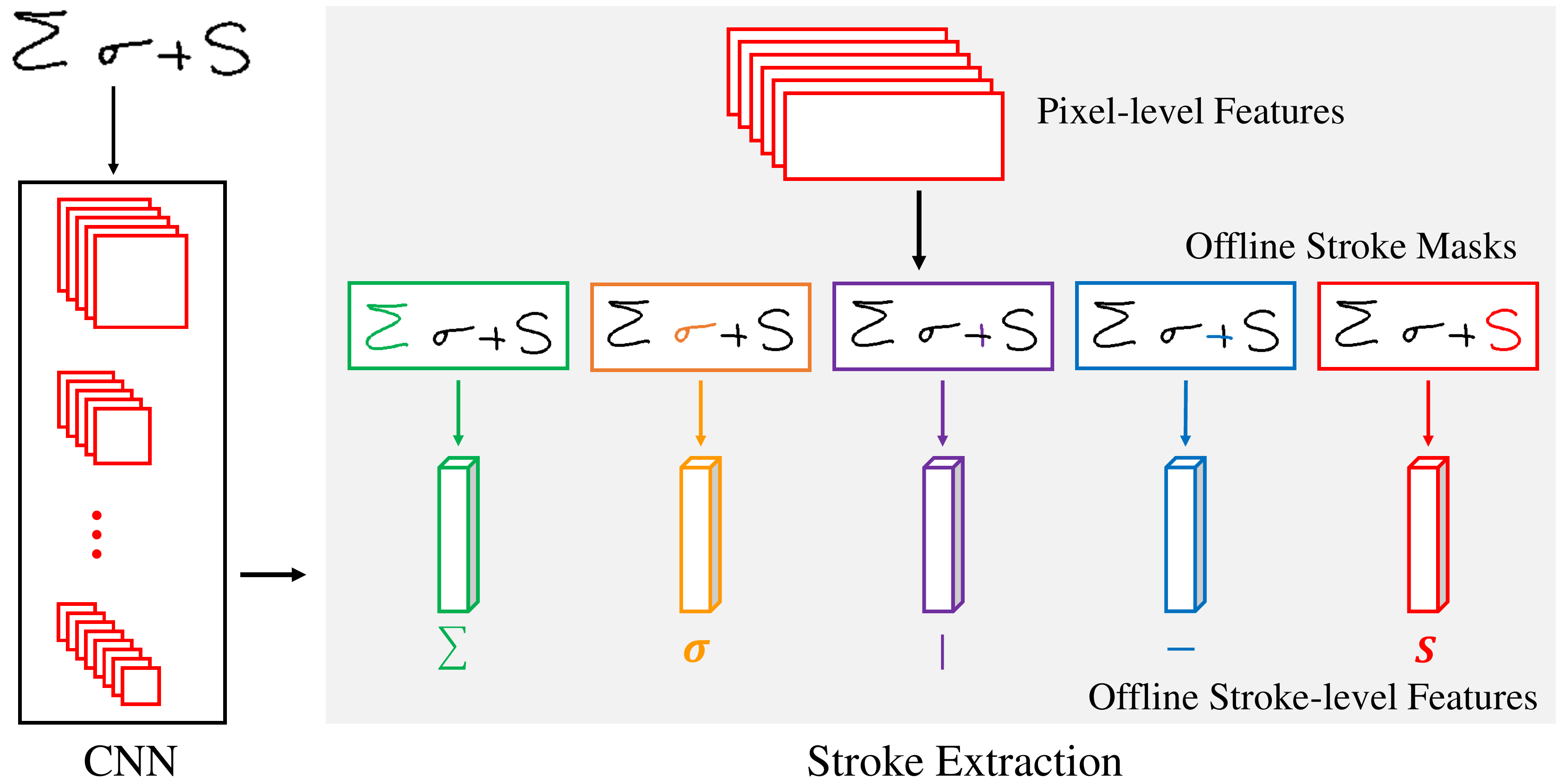}}
\caption{The architecture of offline encoder. The left part is pixel-level feature extraction from input images using a deep CNN. The right part is offline stroke-level feature extraction from pixel-level features.}
\label{offline-mask}
\end{figure}

The offline encoder is designed to extract the offline stroke-level features based on $\mathbf{X}^{\text{off}}$ and $\mathbf{Mask}^{\text{off}}$. We first introduce a deeper CNN to extract pixel-level features, which is illustrated in the left part of Figure~\ref{offline-mask}. The output of CNN encoder is a tensor of size $H \times W \times D$. Note that ${H}_{\text{in}}$ and ${W}_{\text{in}}$ are multiples of $H$ and $W$ based on the number of downsampling in CNN encoder, respectively. We transform this tensor into a variable-length vector sequence $\mathbf{B}=\left \{ \mathbf{b}_{1},\mathbf{b}_{2},\cdots, \mathbf{b}_{H \times W} \right \}$ as the pixel-level features and each element is a $D$-dimensional vector.

Similar to the online encoder, we utilize offline stroke masks to convert pixel-level features into offline stroke-level features, which is illustrated in the right part of Figure~\ref{offline-mask}. First, the same number of downsampling as that in CNN encoder is used to process offline stroke masks, which converts each offline stroke mask from a matrix $\mathbf{mask}^{\text{off}}_{j}$ of size ${H}_{\text{in}} \times {W}_{\text{in}}$ to a matrix $\mathbf{pmask}^{\text{off}}_{j}$ of size $H \times W$. Then we transform each offline mask into a $(H \times W) $-dimensional vector and offline stroke-level features are extracted from pixel-level features as:
\begin{equation}
\mathbf{s}^{\text{off}}_{j}= \frac{\mathbf{pmask}^{\text{off}}_{j} }{||\mathbf{pmask}^{\text{off}}_{j}||_1} \mathbf{B} \quad \quad \mathbf{S}^{\text{off}}=\left \{ \mathbf{s}^{\text{off}}_{1},\mathbf{s}^{\text{off}}_{2},\cdots, \mathbf{s}^{\text{off}}_{M} \right \}
\end{equation}
where $\mathbf{s}^{\text{off}}_{j}$ is a $D$-dimension vector and $\mathbf{S}^{\text{off}}$ is the final output of offline encoder.

\subsection{Decoder with Attention}
\label{sec:decoder}
\begin{figure}[htb]
\centerline{\includegraphics[width=0.7\linewidth]{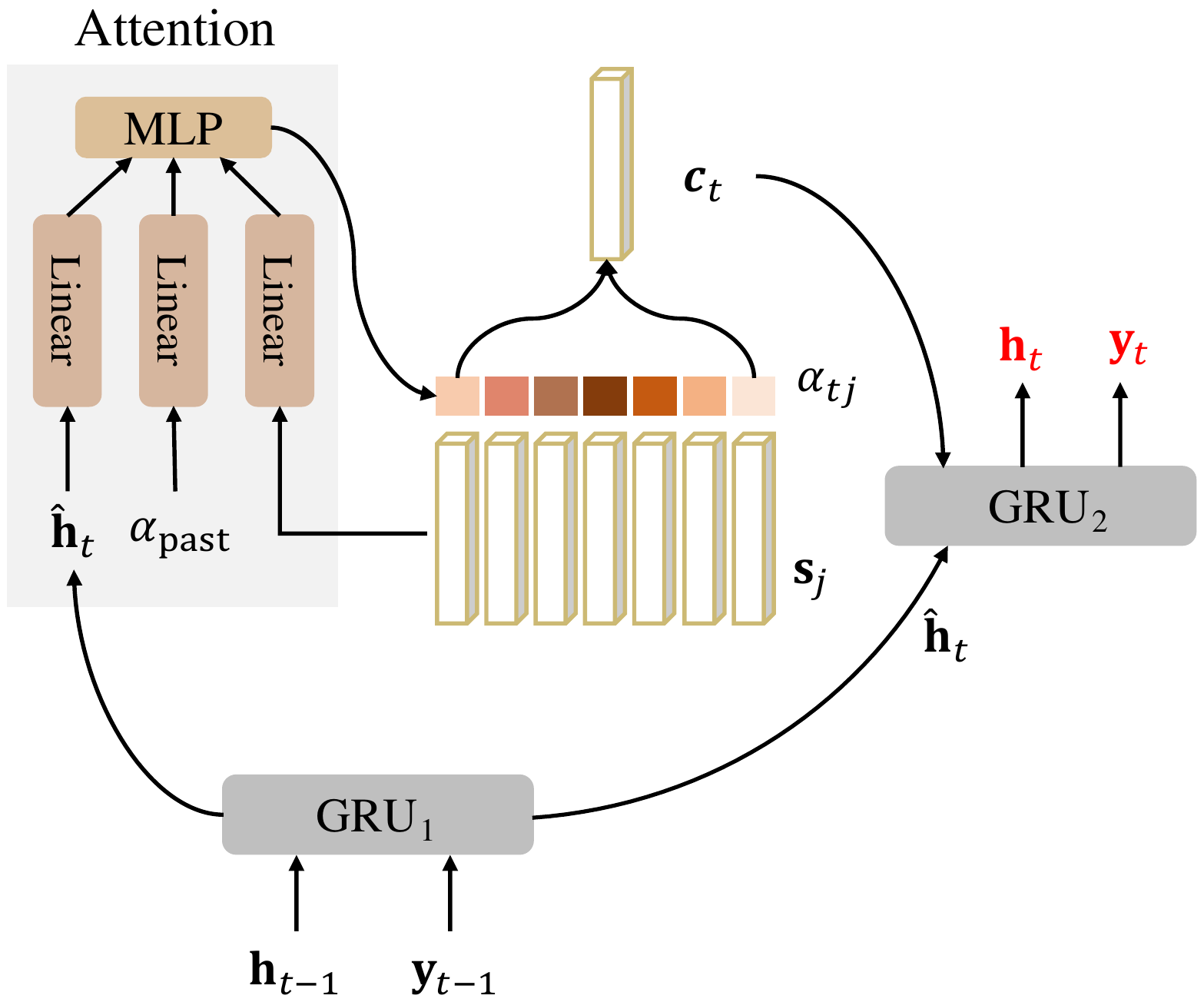}}
\caption{The decoder architecture with two GRU layers and a coverage-based attention. $\alpha_{\text{past}}$ denotes $\sum\nolimits_{\tau=1}^{t-1}\bm{\alpha}_{\tau}$.}
\label{decoder}
\end{figure}

As online and offline stroke-level features ($\mathbf{S}^{\text{on}}$ and $\mathbf{S}^{\text{off}}$) are both vector sequences, we employ the same decoder architecture with a coverage-based attention for both online and offline SCAN. But the parameters contained in decoder and attention are not shared. As shown in Figure~\ref{decoder}, the decoder accepts online or offline stroke-level features and generates a LaTeX sequence for recognition:
\begin{equation}
\mathbf{Y}=\left \{ \mathbf{y}_{1},\mathbf{y}_{2},\cdots, \mathbf{y}_{C} \right \},\mathbf{y}_{i}\in  \mathbb{R}^{K}
\end{equation}
where $K$ is the number of total math symbols in the vocabulary and $C$ is the length of LaTeX sequence. To address the problem that the stroke-level features have a variable length and the length of LaTeX string is not fixed, we employ an intermediate fixed-size vector $\mathbf{c}_{t}$, namely context vector generated by a unidirectional GRU with a coverage-based attention, which will be described later. Then another unidirectional GRU is adopted to produce the LaTeX sequence symbol by symbol. The decoder structure can be denoted as:
\begin{align}
& \mathbf{\hat{h}}_{t}=\textrm{GRU}_{1}\left ( \mathbf{y}_{t-1},\mathbf{h}_{t-1} \right ) \label{eqn:first GRU} \\
& \mathbf{c}_{t}=f_{\text{att}}\left (\mathbf{\hat{h}}_{t},\mathbf{S} \right ) \label{eqn:att} \\
& \mathbf{h}_{t}=\textrm{GRU}_{2}\left ( \mathbf{c}_{t},\mathbf{\hat{h}}_{t} \right ) \label{eqn:second GRU}
\end{align}
where $\textrm{GRU}_{1}$, $\textrm{GRU}_{2}$ indicate two GRU layers, $f_{\text{att}}$ denotes the coverage-based attention, $\mathbf{\hat{h}}_{t}$ and $\mathbf{h}_{t}$ represent the hidden states of the first and the second GRU layers, $\mathbf{S}$ denotes online or offline stroke-level features.
Besides, we utilize $\mathbf{\hat{h}}_{t}$ instead of $\mathbf{h}_{t-1}$ to calculate attention coefficients as we believe that $\mathbf{\hat{h}}_{t}$ is a more accurate representation of the current alignment information than $\mathbf{h}_{t-1}$.

The probability of each predicted symbol is then computed by the context vector $\mathbf{c}_{t}$, the hidden state of the second GRU layer $\mathbf{h}_{t}$ and one-hot vector of previous output symbol $\mathbf{y}_{t-1}$ using the following equation:
\begin{equation}
\label{probability}
p\left ( \mathbf{y}_{t} \right )=g\left ( \mathbf{W}_{o}\phi \left ( \mathbf{E}\mathbf{y}_{t-1}+\mathbf{W}_{h}\mathbf{h}_{t}+\mathbf{W}_{c}\mathbf{c}_{t} \right ) \right )
\end{equation}
where $g$ represents the softmax activation function and $\phi$ represents the maxout activation function. $\mathbf{W}_{o}\in \mathbb{R}^{K\times \frac{m}{2}}$, $\mathbf{W}_{h}\in \mathbb{R}^{m\times n}$, $\mathbf{W}_{c}\in \mathbb{R}^{m\times D}$, and $\mathbf{E} \in \mathbb{R}^{m\times K}$ denotes the embedding matrix. $m$ and $n$ are dimensions of embedding and GRU decoder.

Attention mechanism is widely adopted in sequence learning~\cite{attention1,attention2,attention3}. It is intuitive that for each predicted symbol, only parts of the input rather than the entire input is necessary to provide the useful information, which means only a subset of feature vectors mainly contribute to the recognition. As shown in Figure~\ref{decoder}, we introduce a coverage-based attention, ${f}_{\text{att}}$, which can be represented as:
\begin{align}
& \mathbf{F}=\mathbf{Q}\ast\sum\nolimits_{\tau=1}^{t-1}\bm{\alpha}_{\tau} \\
& {e}_{tj}=\bm{\nu}^{\rm T}_{\text{att}}\tanh\left ( \mathbf{W}_{\text{att}}\mathbf{\hat{h}}_{t}+ \mathbf{U}_{\text{att}} \mathbf{s}_{j}+ \mathbf{U}_{f}\mathbf{f}_{j} \right )
\end{align}
 where ${e}_{tj}$ denotes the energy of stroke-level feature vector $\mathbf{s}_{j}$ in decoding step $t$. $\mathbf{F}$ with its element $\mathbf{f}_{j}$ as the coverage vector is computed by feeding the past attention into a convolution layer $\mathbf{Q}$ with $q$ output channels, which can help alleviate the problem of standard attention mechanism, namely lack of coverage~\cite{coverage}. Let $n'$ denotes the dimension of the attention, then $\bm{\nu}_{\text{att}} \in {\mathbb{R}^{{n'}}}$, ${{\mathbf{W}}_{\text{att}}} \in {\mathbb{R}^{{n'} \times n}}$, ${{\mathbf{U}}_{\text{att}}} \in {\mathbb{R}^{{n'} \times D}}$, $\mathbf{U}_{f} \in {\mathbb{R}^{{n'} \times q}}$.

The attention coefficients ${\alpha}_{tj}$ can be obtained by feeding ${e}_{tj}$ into a softmax function, which is utilized to calculate the context vector as:
\begin{equation}
{\alpha}_{tj}=\frac{\exp\left({e}_{tj}\right)}{\sum\nolimits_{k=1}^{M}{\exp \left( {e}_{tk} \right)}} \quad \quad \mathbf{c}_{t}=\sum_{j=1}^{M}{\alpha} _{tj}\mathbf{s}_{j}
\end{equation}

\subsection{Stroke-level Attention Guider}\label{stroke_att_guider}
 For online HMER, the correspondence information between strokes and symbols is provided in the training stage. For example, there is an expression ``$s + 2$'' which consists of four strokes: the first stroke for ``$s$'', the second and the third strokes for ``$+$'' and the last stroke for ``$2$''. Obviously, when we predict the symbol ``$+$'', the coverage-based attention should be supposed to only attend the second and the third strokes. Generally for the symbol ${w}_{t}$ in time step $t$, we first introduce an oracle attention map, $\bm{\gamma}_{t}=\left \{ {\gamma}_{tj} | j=1,2,\ldots,M \right \}$ with $\gamma_{tj} = \frac{1}{M'}$ if the ${j}^{\text{th}}$ stroke belongs to the symbol ${w}_{t}$, otherwise 0, where $M$ denotes the number of all strokes and $M'$ denotes the number of strokes belonging to the symbol ${w}_{t}$. We can regard $\gamma_{tj}$ and $\alpha_{tj}$ as two probability distributions because $\sum\nolimits_{j=1}^{M}\gamma_{tj}=\sum\nolimits_{j=1}^{M}\alpha_{tj}=1$ and it is intuitive to employ the cross entropy function as the stroke-level attention guider:
\begin{equation}\label{guider}
{G}_{t}= - \sum\nolimits_{j=1}^{M} \gamma_{tj} \log \alpha_{tj}
\end{equation}
Note that for spatial structure, such as ``$\wedge$", ``$\{$'' and  ``$\}$'', which are used to meet the requirement of LaTeX grammar, we simply remove the guider as they are lack of explicit alignments to strokes. This stroke-level attention guider is adopted as a regularization item for parameter learning as elaborated in Section~\ref{training_procedure}.

\section{Multi-modal SCAN}
In this section, we discuss multi-modal SCAN, which can take both advantages of online and offline modalities for online HMER. First, we employ a multi-modal encoder with both online and offline encoder to extract online stroke-level features $\mathbf{S}^{\text{on}}$ and offline stroke-level features $\mathbf{S}^{\text{off}}$, as shown in Section~\ref{sec:online encoder} and Section~\ref{sec:offline encoder}. Then two fusion strategies are proposed for multi-modal SCAN, namely the decoder fusion (denoted as MMSCAN-D) and the encoder fusion (denoted as MMSCAN-E). In the decoder fusion, similar to our previous work~\cite{Wang2019MAN}, a multi-modal attention equipped with re-attention mechanism to fuse online and offline stroke-level features is introduced. More importantly, SCAN makes the fusion of online and offline stroke-level features in encoder become possible as it provides oracle alignments between online and offline modalities.

\subsection{Decoder Fusion}

\begin{figure}[htb]
\centerline{\includegraphics[width=0.95\linewidth]{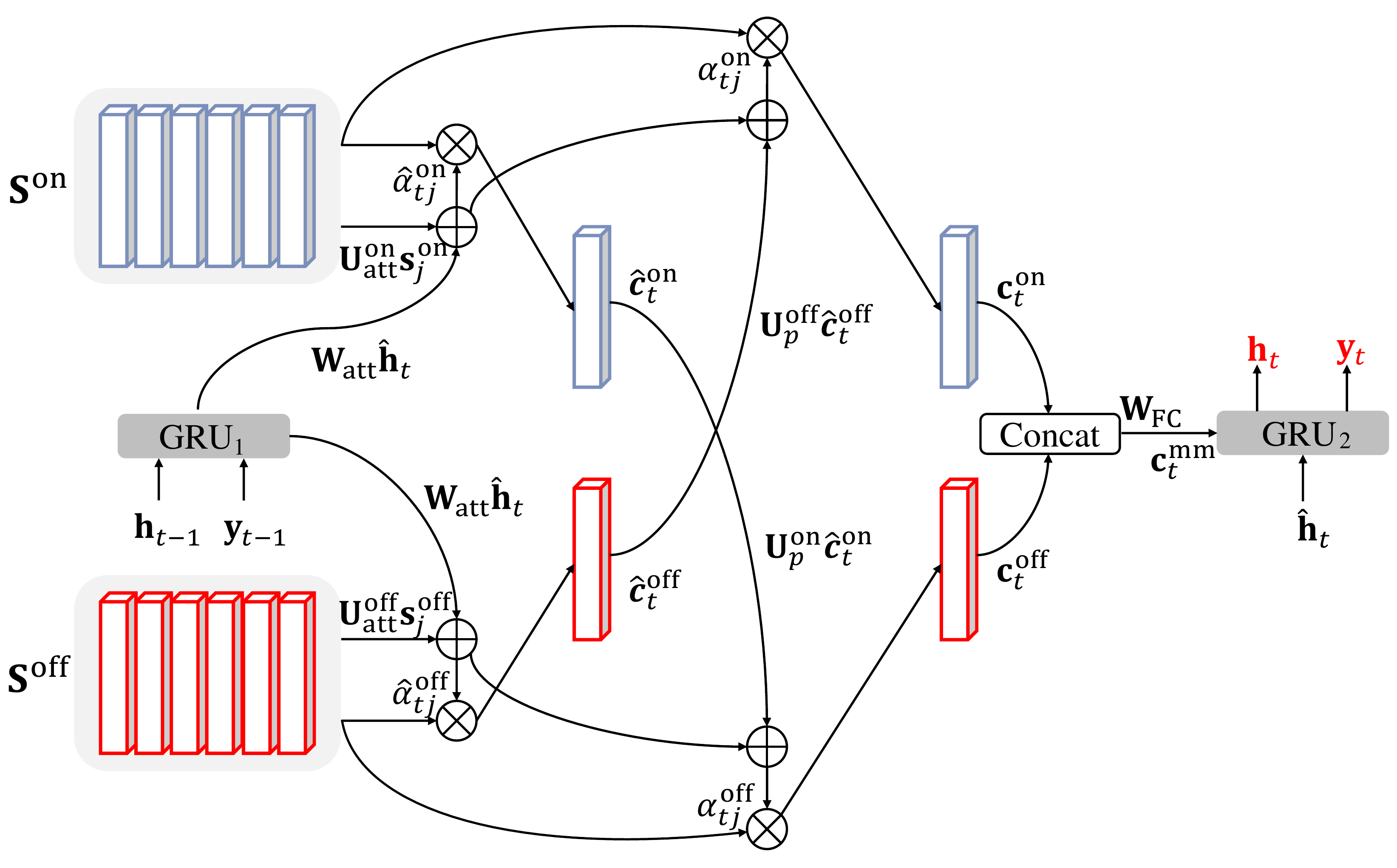}}
\caption{The two-stage re-attention mechanism with pre-attention and fine-attention models. To simplify the illustration, we have omitted the coverage vectors and activation functions.}
\label{re-attention}
\end{figure}

To fully utilize the complementarities between online and offline modalities, a two-stage re-attention mechanism is designed with pre-attention and fine-attention models, which is illustrated in Figure~\ref{re-attention}. Actually the decoder structure here is similar to the single-modal case as described in Eq.~\labelcref{eqn:first GRU,eqn:att,eqn:second GRU}. The main difference is that ${f}_{\text{att}}$ in Eq.~(\ref{eqn:att}) is replaced by the re-attention mechanism, which accepts online and offline stroke-level features and generates a multi-modal stroke-level context vector, $\mathbf{c}^{\text{mm}}_{t}$. In the first stage, the pre-attention model can be represented as:
\begin{equation}\label{df_preatt}
\mathbf{\hat{c}}_{t}^{\text{on}}={f}^{\text{on}}_{\text{att}}\left( \mathbf{\hat{h}}_{t},\mathbf{S}^{\text{on}}\right) \quad \quad
\mathbf{\hat{c}}_{t}^{\text{off}}={f}^{\text{off}}_{\text{att}}\left( \mathbf{\hat{h}}_{t},\mathbf{S}^{\text{off}}\right)
\end{equation}
where $\mathbf{\hat{c}}_{t}^{\text{on}}$ and $\mathbf{\hat{c}}_{t}^{\text{off}}$ denote two single-modal stroke-level context vectors. Note that the superscripts ``$\text{on}$'' and ``$\text{off}$'' in Eq.~(\ref{df_preatt}) are only used to distinguish coverage-based attention ${f}_{\text{att}}$ over online and offline stroke-level features as the attention parameters are not shared.

Based on the results of the pre-attention model, the fine-attention model is employed to generate multi-modal stroke-level context vector $\mathbf{c}^{\text{mm}}_{t}$ in the second stage. Compared with the pre-attention model, the fine-attention model adds the context vector of one modality from the pre-attention model as the auxiliary information to improve the attention of another modality, which is implemented as:
\begin{equation}
{\alpha}_{tj}^{\text{on}}=g\left ( \bm{\nu}_{\text{att}}^{\rm T}\tanh\left ( \mathbf{W}_{\text{att}}\mathbf{\hat{h}}_{t}+ \mathbf{U}_{\text{att}}^{\text{on}} \mathbf{s}^{\text{on}}_{j} + \mathbf{U}_{f}^{\text{on}}\mathbf{f}_{j}^{\text{on}}+\mathbf{U}_{p}^{\text{off}}\mathbf{\hat{c}}_{t}^{\text{off}} \right ) \right )
\end{equation}
\begin{equation}
{\alpha}_{tj}^{\text{off}}=g\left ( \bm{\nu}_{\text{att}}^{\rm T}\tanh\left (\mathbf{W}_{\text{att}}\mathbf{\hat{h}}_{t}+ \mathbf{U}_{\text{att}}^{\text{off}} \mathbf{s}^{\text{off}}_{j} + \mathbf{U}_{f}^{\text{off}}\mathbf{f}_{j}^{\text{off}}+\mathbf{U}_{p}^{\text{on}}\mathbf{\hat{c}}_{t}^{\text{on}} \right ) \right )
\end{equation}
where $\mathbf{U}_{p}^{\text{on}}\in \mathbb{R}^{n'\times D}$, $\mathbf{U}_{p}^{\text{off}}\in \mathbb{R}^{n'\times D}$. Then the stroke-level context vectors of fine-attention model are calculated as:
\begin{equation}
\mathbf{c}_{t}^{\text{on}}= \sum\nolimits_{j=1}^{M}{\alpha}_{tj}^{\text{on}}\mathbf{s}^{\text{on}}_{j} \quad \quad
\mathbf{c}_{t}^{\text{off}}= \sum\nolimits_{j=1}^{M}{\alpha}_{tj}^{\text{off}}\mathbf{s}^{\text{off}}_{j}
\end{equation}
Finally the multi-modal stroke-level context vector $\mathbf{c}^{\text{mm}}_{t}$ can be obtained as:
\begin{equation}
\mathbf{c}^{\text{mm}}_{t}=\tanh \left( \mathbf{W}_{\text{FC}}  \begin{bmatrix}
\mathbf{c}_{t}^{\text{on}}\\
\mathbf{c}_{t}^{\text{off}}
\end{bmatrix} \right)
\end{equation}
where $\mathbf{W}_{\text{FC}} \in \mathbb{R}^{D\times 2D}$.

The re-attention can be also equipped with stroke-level attention guider as described in Section~\ref{stroke_att_guider} and the main difference is that here we utilize oracle attention map $\bm{\gamma}_{t}$ to supervise the learning of both online and offline attention coefficients as:
\begin{equation}
{G}_{t}= - \left(\sum\nolimits_{j=1}^{M} \gamma_{tj} \log \alpha^{\text{on}}_{tj} + \sum\nolimits_{j=1}^{M} \gamma_{tj} \log \alpha^{\text{off}}_{tj}\right)
\end{equation}

\subsection{Encoder Fusion}
A key component of multi-modal learning is to fuse features from different modalities. In our previous work~\cite{Wang2019MAN}, point-level and pixel-level features are extracted from the inputs of online and offline modalities. On account of the problem that these two types of features are unaligned, we can only fuse online and offline modalities in decoder.

\begin{figure*}[htb]
\centerline{\includegraphics[width=0.98\linewidth]{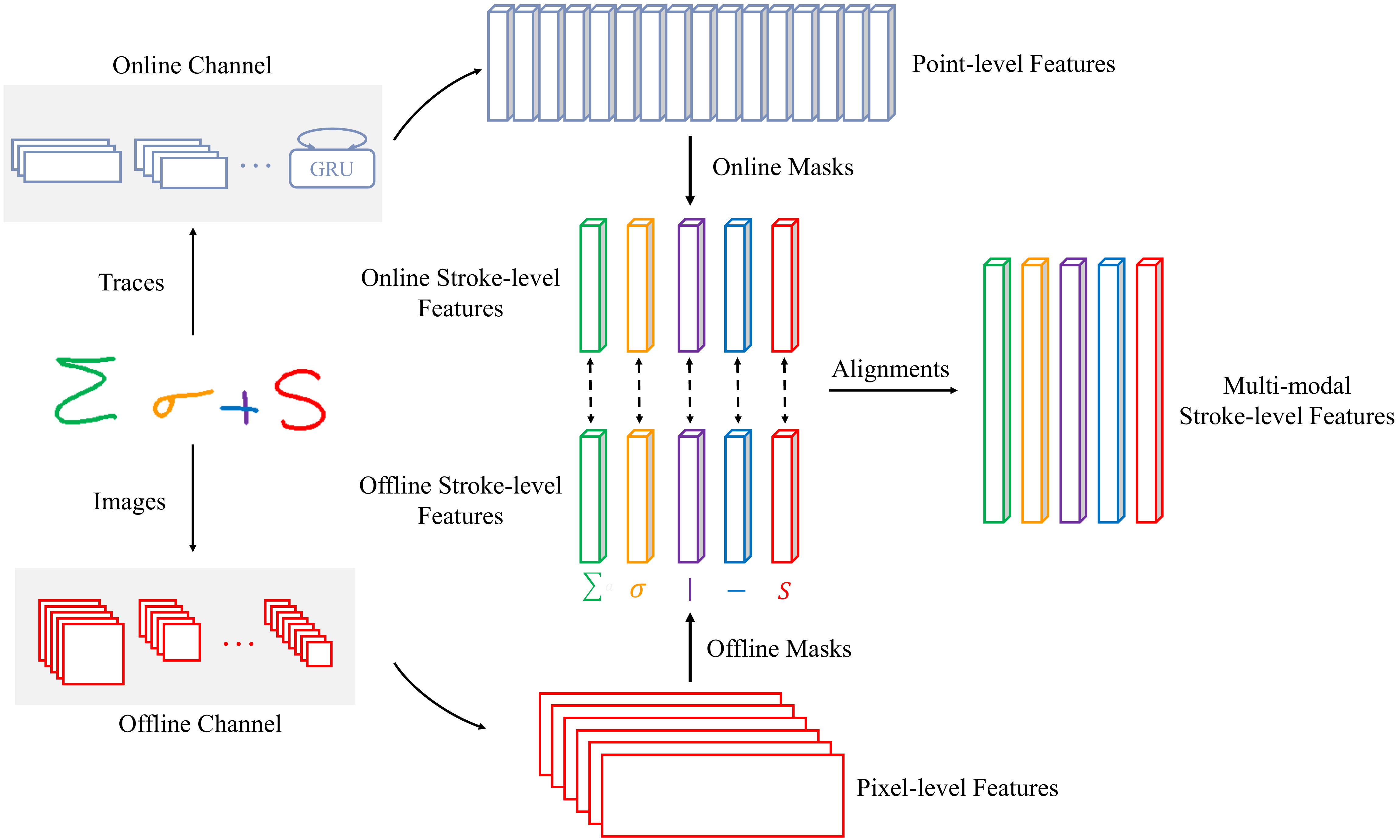}}
\caption{Encoder fusion to generate multi-modal stroke-level features with the oracle alignments between online and offline stroke-level features.}
\label{alignments}
\end{figure*}

However, as illustrated in Figure~\ref{alignments}, SCAN converts point-level and pixel-level features into online and offline stroke-level features. Inherently, there are oracle alignments between online and offline modalities in terms of stroke-level features. Specifically, online and offline stroke-level features are one-to-one correspondence, both indicating the high-level representations of a certain stroke. Therefore, we can fuse online and offline stroke-level features into multi-modal stroke-level features as:
\begin{equation}
\mathbf{S}^{\text{mm}}=\left\{ \mathbf{s}^{\text{mm}}_{1}, \mathbf{s}^{\text{mm}}_{2}, \ldots ,\mathbf{s}^{\text{mm}}_{M} \right\} \quad \quad \mathbf{s}^{\text{mm}}_{j}=
\begin{bmatrix}
\mathbf{s}^{\text{on}}_{j}\\
\mathbf{s}^{\text{off}}_{j}
\end{bmatrix}
\end{equation}

With the multi-modal stroke-level features, we employ a decoder with coverage-based attention and the stroke-level attention guider described in \ref{stroke_att_guider} to generate the recognition result. The structure is similar to that illustrated in Section \ref{sec:decoder} by replacing $\mathbf{S}^{\text{on}}$/$\mathbf{S}^{\text{off}}$ with $\mathbf{S}^{\text{mm}}$, which can be denoted as:
\begin{align}
& \mathbf{\hat{h}}_{t}=\textrm{GRU}_{1}\left ( \mathbf{y}_{t-1},\mathbf{h}_{t-1} \right )  \\
& \mathbf{c}^{\text{mm}}_{t}=f_{\text{att}}\left (\mathbf{\hat{h}}_{t},\mathbf{S}^{\text{mm}} \right )  \\
& \mathbf{h}_{t}=\textrm{GRU}_{2}\left ( \mathbf{c}^{\text{mm}}_{t},\mathbf{\hat{h}}_{t} \right )
\end{align}

By comparison of MMSCAN-E and MMSCAN-D, although MMSCAN-D introduces re-attention to help information interaction between two modalities, it still only considers information from one single modality in the pre-attention model. Moreover, the errors in the pre-attention model will be inherited by the fine-attention model which might degrade the performance. Nevertheless, MMSCAN-E makes full use of stroke constrained information to obtain oracle alignments between online and offline stroke-level features and fuse them in encoder. Consequently, MMSCAN-E takes the fusion one step before MMSCAN-D, which can potentially improve the recognition performance of HMER.

\section{Training and Testing Procedures}
\subsection{Training}
\label{training_procedure}
Our models aim to maximize the predicted symbol probability as shown in Eq.~(\ref{probability}) and employ cross entropy (CE) as the criterion. The objective function for optimization, which consists of CE criterion and the stroke-level attention guider, is shown as follows:
\begin{equation}\label{objective}
O = - \sum\nolimits_{t=1}^C \log p({w_t}|{\mathbf{y}_{t-1},\mathbf{X^{\text{on}}},\mathbf{X^{\text{off}}}}) + \lambda \sum\nolimits_{t=1}^C G_t
\end{equation}
where $w_t$ represents the ground truth word at time step $t$, $C$ is the length of output string in LaTeX format, $G_t$ is the stroke-level attention guider, and $\lambda$ is set to 0.2. Note that for single-modal HMER, only one of $\mathbf{X^{\text{on}}}$ and $\mathbf{X^{\text{off}}}$ is used. Besides, we set weight decay to $10^{-5}$ to reduce overfitting.

\begin{figure}[htb]
\centerline{\includegraphics[width=0.7\linewidth]{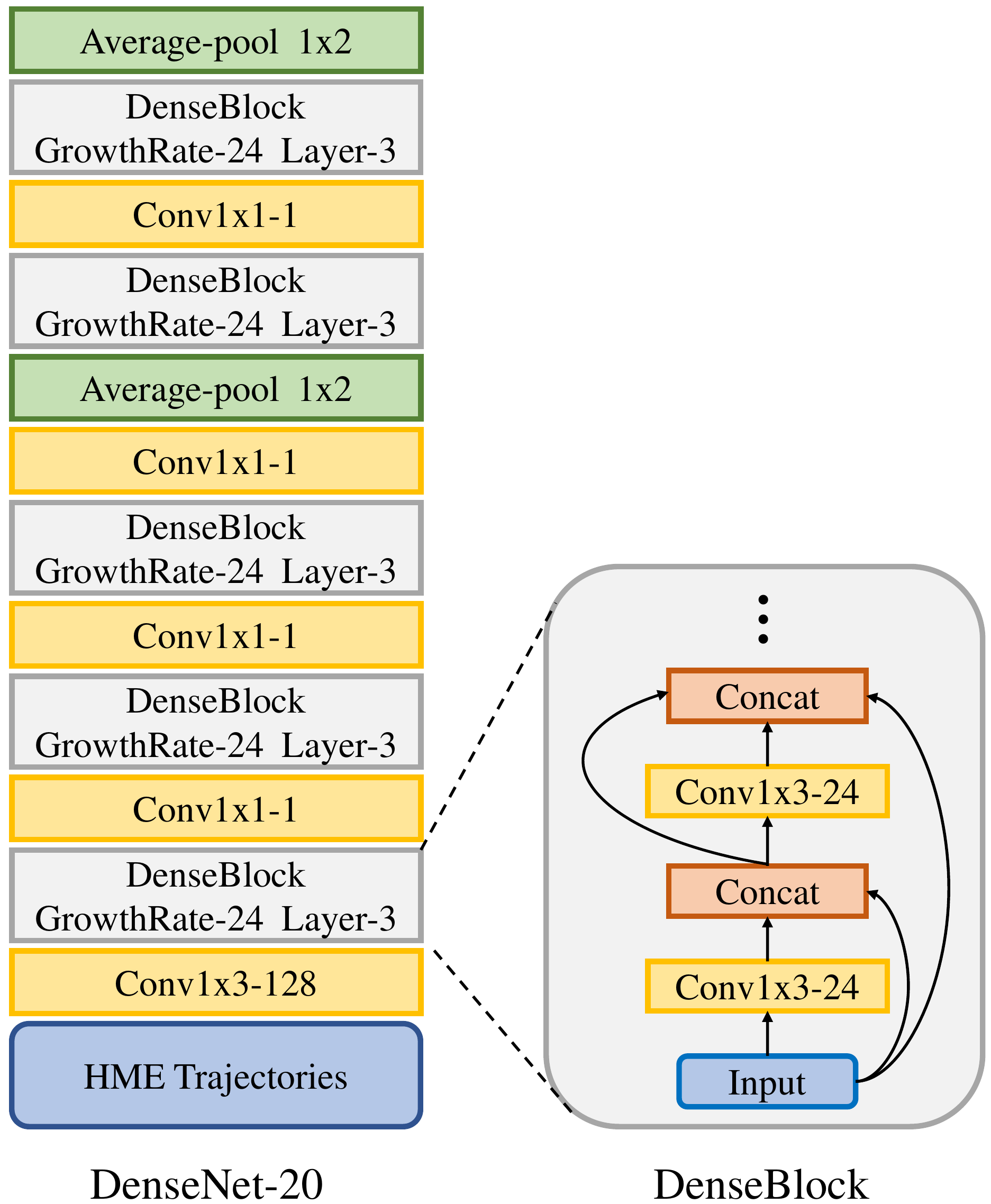}}
\caption{The architecture of DenseNet-20.}
\label{DenseNet20}
\end{figure}

There are three kinds of encoders in this study, namely online encoder, offline encoder and multi-modal encoder while multi-modal encoder is the combination of online encoder and offline encoder. The online encoder is a CNN-GRU architecture. The CNN part is a DenseNet as illustrated in Figure~\ref{DenseNet20}, with 5 dense blocks in the main branch. $1 \times 2$ average pooling is applied after the third and fifth dense blocks, which reduces the length of input point sequence by a factor of 4. The growth rate is set to 24 and the compression factor in transition layer is set to 1. As shown in the right part of Figure~\ref{DenseNet20}, each dense block without bottleneck structure has 3 convolutional layers with kernel size $1  \times 3$ and 24 output channels. The GRU part is two layers of bidirectional GRU and each GRU layer has 250 forward and 250 backward units.

\begin{figure}[htb]
\centerline{\includegraphics[width=0.7\linewidth]{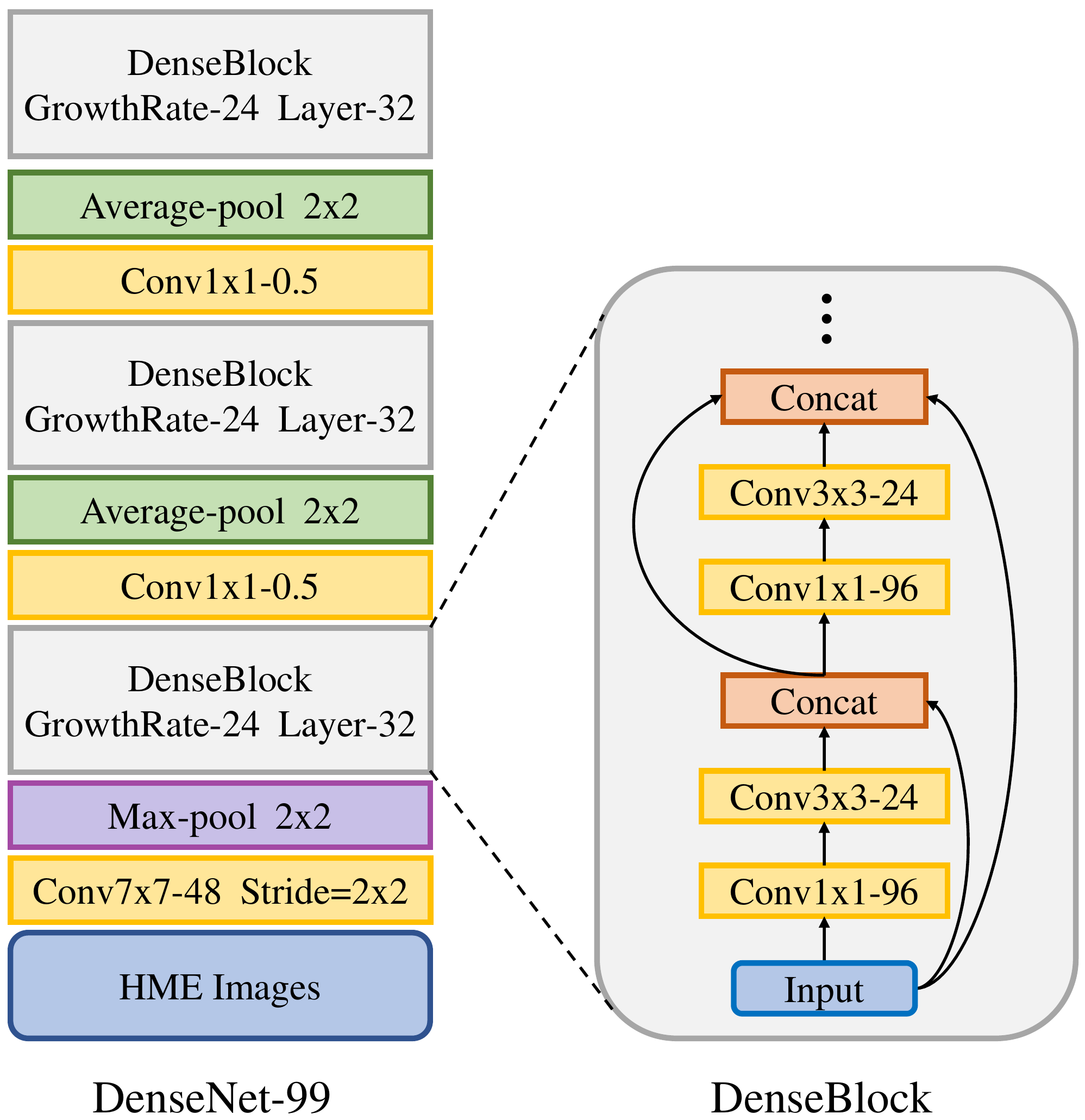}}
\caption{The architecture of DenseNet-99.}
\label{DenseNet99}
\end{figure}

The offline encoder is a deeper DenseNet as illustrated in Figure~\ref{DenseNet99}, with 3 dense blocks in the main branch. $1 \times 1 $ convolution followed by $ 2 \times 2$ average pooling between every two contiguous dense blocks is used. The growth rate is set to 24 and the compression factor in transition layer is set to 0.5. As shown in the right part of Figure~\ref{DenseNet99}, each dense block adopts the bottleneck structure, i.e., a $1 \times 1$ convolution is introduced before each $3 \times 3$ convolution to reduce the input to 96 feature maps and the total number of convolutional layers in each block is 32. Note that there are additional fully connected layers on top of online and offline channels of multi-modal encoder to convert the output dimensions of these two channels to be the same, namely $D=500$.

The decoder adopts 2 unidirectional GRU layers and each layer has 256 forward GRU units. The embedding dimension $m$ and GRU decoder dimension $n$ are both set to 256 while the attention dimension $n'$ is 500. The kernel sizes of convolution layers $\mathbf{Q}$ are set to $1 \times 7$ for online modality and $11 \times 11$ for offline modality. We train our model by the adadelta algorithm~\cite{Adadelta} for optimization and the corresponding hyperparameters are set as $\rho =0.95$, $\varepsilon =10^{-8}$.
\subsection{Testing}
In the recognition stage, we expect to obtain the most likely LaTeX string as:
\begin{equation}\label{decoding target}
{\bf{\hat y}} = \mathop {\arg \max }\limits_{\bf{y}} \log P\left( {{\bf{y}}|{\bf{x}}} \right)
\end{equation}
Different from the training stage, we do not have the ground truth of the previous predicted symbol. Consequently, we employ a simple left-to-right beam search algorithm~\cite{BeamSearch} to implement the decoding procedure, beginning with the start-of-sentence token $<\!sos\!>$. At each time step, we maintain a set of 10 partial hypotheses. Each hypothesis is expanded with every possible symbol and only the hypotheses with 10 minimal scores are kept:
\begin{equation}\label{score}
\mathbf{S}_t = \mathbf{S}_{t-1} - \log p({\mathbf{y}_t}|{\mathbf{y}_{t-1},\mathbf{x}})
\end{equation}
where $\mathbf{S}_{t-1}$ and $\mathbf{S}_{t}$ represent the scores at time steps $t-1$ and $t$, respectively. $p({\mathbf{y}_t}|{\mathbf{y}_{t-1},\mathbf{x}})$ denotes the probability of all predicted symbols in the dictionary. The prediction procedure for each hypothesis ends when the output symbol meets the end-of-sentence token $<\!eos\!>$.

\section{Experiments}
In this section, we design a set of experiments to validate the effectiveness of the proposed SCAN by answering the following questions:
\begin{description}
\item[Q1] Is the proposed single-modal SCAN effective for online HMER?
\item[Q2] Is the proposed multi-modal SCAN using encode/decoder fusion effective?
\item[Q3] How does SCAN improve the performance by attention visualization?
\item[Q4] Can SCAN help accelerate the recognition speed?
\end{description}
The experiments are all implemented with Pytorch 0.4.1~\cite{pytorch} and an NVIDIA GeForce GTX 1080Ti 11G GPU. And our source code will be publicly available.

\subsection{Dataset and Metric}
Our experiments are conducted on CROHME competition database~\cite{crohme14,crohme16}, which is currently the most widely used dataset for HMER. The CROHME 2014 competition dataset consists of a training set of 8836 HMEs and a testing set of 986 HMEs. The CROHME 2016 competition dataset only includes a testing set of 1147 HMEs. There are totally 101 math symbol classes and none of the handwritten expressions in the testing set appears in the training set. We apply CROHME 2014 training set as our training set and evaluate the performance of our models on CROHME 2014 testing set and CROHME 2016 testing set. Besides, we also evaluate our models on the latest CROHME 2019 competition dataset of 1199 HMEs.

The main metric in this study is expression recognition rate (ExpRate)~\cite{crohmeDatabase}, i.e., the percentage of predicted mathematical expressions matching the ground truth.
Besides, we list the structure recognition rate (StruRate)~\cite{crohmeDatabase}, which only focuses on whether the structure is correctly recognized and ignores symbol recognition errors.

\subsection{Evaluation of Single-modal SCAN (Q1)}
\label{sec:singleModalityExperiment}

\begin{table}[htb]
\caption{\label{tab:onlinePerformance}{Performance comparison of different encoder-decoder approaches for the online modality using point-level features (TAP), online stroke-level features (OnSCAN), and feature fusion in decoder (OnSCAN+TAP) on CROHME 2014 and CROHME 2016 testing sets.}}
\centering
\begin{tabular}{c c c c c}
\toprule
\multirow{2}{*}{\textbf{System}} & \multicolumn{2}{c}{\textbf{CROHME 2014}} & \multicolumn{2}{c}{\textbf{CROHME 2016}} \\
\cmidrule(lr){2-3}
\cmidrule(lr){4-5}
 & \textbf{ExpRate}  & \textbf{StruRate} & \textbf{ExpRate}  & \textbf{StruRate}\\
\midrule
TAP \cite{Wang2019MAN} & 48.47\%  & 67.24\% & 44.81\%  & 63.12\% \\
OnSCAN & 51.22\%  & 70.49\% & 46.12\%  & 65.30\% \\
OnSCAN+TAP & 52.64\%  & 70.89\% & 47.17\%  & 66.78\% \\
\bottomrule
\end{tabular}
\end{table}

In this section, we examine the effectiveness of single-modal SCAN. First, we investigate the performance of different encoder-decoder approaches for the online modality as shown in Table~\ref{tab:onlinePerformance}. TAP refers to the improved version of encoder-decoder approach using point-level features as in~\cite{Wang2019MAN}. OnSCAN+TAP denotes the decoder fusion of TAP using point-level features and OnSCAN using online stroke-level features via the multi-modal attention in~\cite{Wang2019MAN}. The ExpRate is increased from 48.47\% to 51.22\% on CROHME 2014 testing set and from 44.81\% to 46.12\% on CROHME 2016 testing set after replacing point-level features (TAP) with online stroke-level features (OnSCAN). By comparing OnSCAN with OnSCAN+TAP, the ExpRate is increased from 51.22\% to 52.64\% on CROHME 2014 testing set and from 46.12\% to 47.17\% on CROHME 2016 testing set. Similar observations could be made for StruRate. All these results demonstrate the superiority of online stroke-level features as a higher-level representation over the point-level features and the complementarity between them.

\begin{table}[htb]
\caption{\label{tab:offlinePerformance}{Performance comparison of different encoder-decoder approaches for the offline modality using pixel-level features (WAP), offline stroke-level features (OffSCAN), and feature fusion in decoder (OffSCAN+WAP) on CROHME 2014 and CROHME 2016 testing sets.}}
\centering
\begin{tabular}{c c c c c}
\toprule
\multirow{2}{*}{\textbf{System}} & \multicolumn{2}{c}{\textbf{CROHME 2014}} & \multicolumn{2}{c}{\textbf{CROHME 2016}} \\
\cmidrule(lr){2-3}
\cmidrule(lr){4-5}
 & \textbf{ExpRate}  & \textbf{StruRate} & \textbf{ExpRate} & \textbf{StruRate}\\
\midrule
WAP \cite{Wang2019MAN}& 48.38\%  & 70.08\% & 46.82\%  & 66.17\% \\
OffSCAN & 47.67\%  & 68.56\% & 46.64\%  & 65.65\% \\
OffSCAN+WAP & 49.39\%  & 71.81\% & 49.60\%  & 68.18\% \\
\bottomrule
\end{tabular}
\end{table}

Then we compare the performance of different encoder-decoder approaches for the offline modality as shown in Table~\ref{tab:offlinePerformance}. WAP refers to the improved version of encoder-decoder approach using pixel-level features as in~\cite{Wang2019MAN}. OffSCAN+WAP denotes the decoder fusion of WAP using pixel-level features and OffSCAN using offline stroke-level features via the multi-modal attention in~\cite{Wang2019MAN}. Compared with WAP, the ExpRate of OffSCAN is slightly decreased from 48.38\% to 47.67\% on CROHME 2014 testing set and from 46.82\% to 46.64\% on CROHME 2016 testing set. This observation is different from that in online modality by the comparison between TAP and OnSCAN. The reason might be that the pooling operation of 2D images in offline modality leads to higher misalignment between each stroke and the corresponding pixels (or points) than that of 1D sequence in online modality. However, significant improvements could be achieved by OffSCAN+WAP over both WAP and OffSCAN, e.g., with ExpRate increasing from 48.38\%/47.67\% to 49.39\% on CROHME 2014 testing set and from 46.82\%/46.64\% to 49.60\% on CROHME 2016 testing set, which indicates the strong complementarity between the offline stroke-level features and pixel-level features.

\begin{table}[htb]
\caption{\label{tab:singleCROHME2019}{Performance comparison of different encoder-decoder approaches for both online and offline modalities on CROHME 2019 testing set. The expression recognition accuracies with one, two and three errors per expression are represented by ``$\leq 1$'', ``$\leq 2$'' and ``$\leq 3$''.}}
\centering
\begin{tabular}{c c c c c c}
\toprule
\textbf{System} & \textbf{ExpRate} & \bm{$\leq 1$} & \bm{$\leq 2$} & \bm{$\leq 3$} & \textbf{StruRate}  \\
\midrule
TAP \cite{Wang2019MAN}& 44.20\% & 58.80\% & 62.72\% & 63.55\% & 63.64\% \\
OnSCAN & 46.46\% & 62.47\% & 66.14\% & 67.14\% & 66.31\% \\
OnSCAN+TAP & 47.62\% & 62.64\% & 67.06\% & 67.72\% & 67.22\% \\
\midrule
WAP \cite{Wang2019MAN}& 48.12\% & 63.47\% & 67.22\% & 67.97\% & 67.97\% \\
OffSCAN & 47.62\% & 63.14\% & 67.06\% & 67.56\% & 67.81\% \\
OffSCAN+WAP & 49.62\% & 66.89\% & 69.97\% & 70.73\% & 70.56\% \\
\bottomrule
\end{tabular}
\end{table}

To further confirm the generalization of single-modality SCAN, we also evaluate on the latest CROHME 2019 competition database as shown in Table~\ref{tab:singleCROHME2019}. For online modality, OnSCAN can achieve better performance than TAP while OnSCAN+TAP can achieve the best performance. As for offline modality, OffSCAN slightly underperforms WAP but OffSCAN+WAP still yields the best performance. All these variation trends on CROHME 2019 testing set are the same as those on CROHME 2014 and 2016 testing sets, which verify the effectiveness of single-modal SCAN for online HMER in both online modality and offline modality.

\subsection{Evaluation of Multi-modal SCAN (Q2)}
\label{sec:multiModalityExperiment}

\begin{table}[htb]
\caption{\label{tab:multiPerformance}{Performance comparison of different multi-modal approaches on CROHME 2014 and CROHME 2016 testing sets.}}
\centering
\begin{tabular}{c c c c c}
\toprule
\multirow{2}{*}{\textbf{System}} & \multicolumn{2}{c}{\textbf{CROHME 2014}} & \multicolumn{2}{c}{\textbf{CROHME 2016}} \\
\cmidrule(lr){2-3}
\cmidrule(lr){4-5}
 & \textbf{ExpRate}  & \textbf{StruRate} & \textbf{ExpRate} & \textbf{StruRate}\\
\midrule
MAN \cite{Wang2019MAN}  & 52.43\%  & 71.60\%  & 49.87\%  & 68.18\% \\
E-MAN \cite{Wang2019MAN}& 54.05\%  & 72.11\%  & 50.56\%  & 67.39\% \\
MMSCAN-D & 55.38\%   & 71.30\%  & 52.22\%  & 68.35\%  \\
MMSCAN-E & 57.20\%   & 73.94\%  & 53.97\%  & 70.62\%  \\
\bottomrule
\end{tabular}
\end{table}

In Table~\ref{tab:multiPerformance}, we show the performance comparison of different multi-modal approaches on CROHME 2014 and CROHME 2016 testing sets. Please note that MAN and E-MAN are our previously proposed work~\cite{Wang2019MAN} using the decoder fusion of point-level features and pixel-level features. And E-MAN is an enhanced version of MAN by adopting the re-attention mechanism. E-MAN can be considered as the decoder fusion of TAP and WAP while MMSCAN-D is the decoder fusion of OnSCAN and OffSCAN.
So from the point-level/pixel-level feature fusion to online/offline stroke-level feature fusion (E-MAN vs. MMSCAN-D), the ExpRate is increased from 54.05\% to 55.38\% on CROHME 2014 testing set and from 50.56\% to 52.22\% on CROHME 2016 testing set, still demonstrating the superiority of treating stroke as the basic modeling unit rather than point or pixel in the multi-modal case. Besides, as SCAN provides oracle alignments between online and offline modalities, MMSCAN-E using the encoder fusion outperforms MMSCAN-D using the decoder fusion, i.e., with the ExpRate increasing from 55.38\% to 57.20\% on CROHME 2014 testing set and from 52.22\% to 53.97\% on CROHME 2016 testing set, which confirms that the early-stage encoder fusion is better than the late-stage decoder fusion in the SCAN framework.

\begin{table}[htb]
\caption{\label{tab:multiCROHME2019}{Performance comparison of different multi-modal approaches on CROHME 2019 testing set. The expression recognition accuracies with one, two and three errors per expression are represented by ``$\leq 1$'', ``$\leq 2$'' and ``$\leq 3$''.}}
\centering
\begin{tabular}{c c c c c c}
\toprule
\textbf{System} & \textbf{ExpRate} & \bm{$\leq 1$} & \bm{$\leq 2$} & \bm{$\leq 3$} & \textbf{StruRate}  \\
\midrule
MAN & 52.21\% & 66.64\% & 69.97\% & 70.39\% & 70.56\% \\
E-MAN & 52.88\% & 67.64\% & 70.81\% & 71.06\% & 71.06\% \\
MMSCAN-D & 53.88\% & 68.31\% & 70.56\% & 71.14\% & 70.98\% \\
MMSCAN-E & 56.21\% & 69.47\% & 71.64\% & 72.06\% & 71.73\% \\
\bottomrule
\end{tabular}
\end{table}

Furthermore, we evaluate the above approaches on CROHME 2019 testing set in Table~\ref{tab:multiCROHME2019} to show that the improvements are significant and stable. Overall, in comparison to single-modal approaches OnSCAN and OffSCAN, the best performing multi-modal approach MMSCAN-E yields large performance gains, e.g., with an absolute ExpRate gain of 8.59\% and an absolute StruRate gain of 3.92\% on CROHME 2019 testing set (OffSCAN in Table~\ref{tab:singleCROHME2019} vs. MMSCAN-E in Table~\ref{tab:multiCROHME2019}).

\begin{table}[htb]
\caption{\label{tab:sotaPerformance}{Overall performance comparison on CROHME 2014 and 2016 testing sets.}}
\centering
\begin{tabular}{c c c c c}
\toprule
\multirow{2}{*}{\textbf{System}} & \multicolumn{2}{c}{\textbf{CROHME 2014}} & \multicolumn{2}{c}{\textbf{CROHME 2016}} \\
\cmidrule(lr){2-3}
\cmidrule(lr){4-5}
 & \textbf{ExpRate}  & \textbf{StruRate} & \textbf{ExpRate} & \textbf{StruRate}\\
\midrule
UPV   & 37.22\%  & -  & -  & - \\
Wiris   & -  & -  & 49.61\%  & - \\
WYGIWYS   & 35.90\%  & -  & -  & - \\
PAL & 39.66\%  & -  & -  & - \\
TAP & 48.47\%  & 67.24\% & 44.81\%  & 63.12\% \\
WAP & 48.38\%  & 70.08\% & 46.82\%  & 66.17\% \\
E-MAN & 54.05\%  & 72.11\%  & 50.56\%  & 67.39\% \\
\midrule
\textbf{MMSCAN-E} & \textbf{57.20\%} & \textbf{73.94\%}  & \textbf{53.97\%}  & \textbf{70.62}\%  \\
\bottomrule
\end{tabular}
\end{table}

Finally, we make a comparison of our best approach MMSCAN-E and other state-of-the-art approaches on both CROHME 2014 and CROHME 2016 testing sets, as shown in Table~\ref{tab:sotaPerformance}. The system UPV denotes the best system in all submitted systems to CROHME 2014 competition while the system Wiris denotes the best system in all submitted systems to CROHME 2016 competition (only using official training dataset) and the details can be seen in \cite{crohme14,crohme16}. The details of WYGIWYS and PAL can refer to~\cite{Deng2017WYGIWYS} and~\cite{wu2018PAL}, respectively. Please note that the results of the end-to-end approaches are not exactly comparable with traditional approaches in the submitted systems to CROHME competitions as the segmentation error is not explicitly considered. Obviously, the proposed MMSCAN-E significantly outperforms other end-to-end approaches with an ExpRate of 57.20\% on CROHME 2014 testing set and an ExpRate of 53.97\% on CROHME 2016 testing set.

\subsection{Attention Visualization (Q3)}
\begin{figure}[htb]
\centerline{\includegraphics[width=0.95\linewidth]{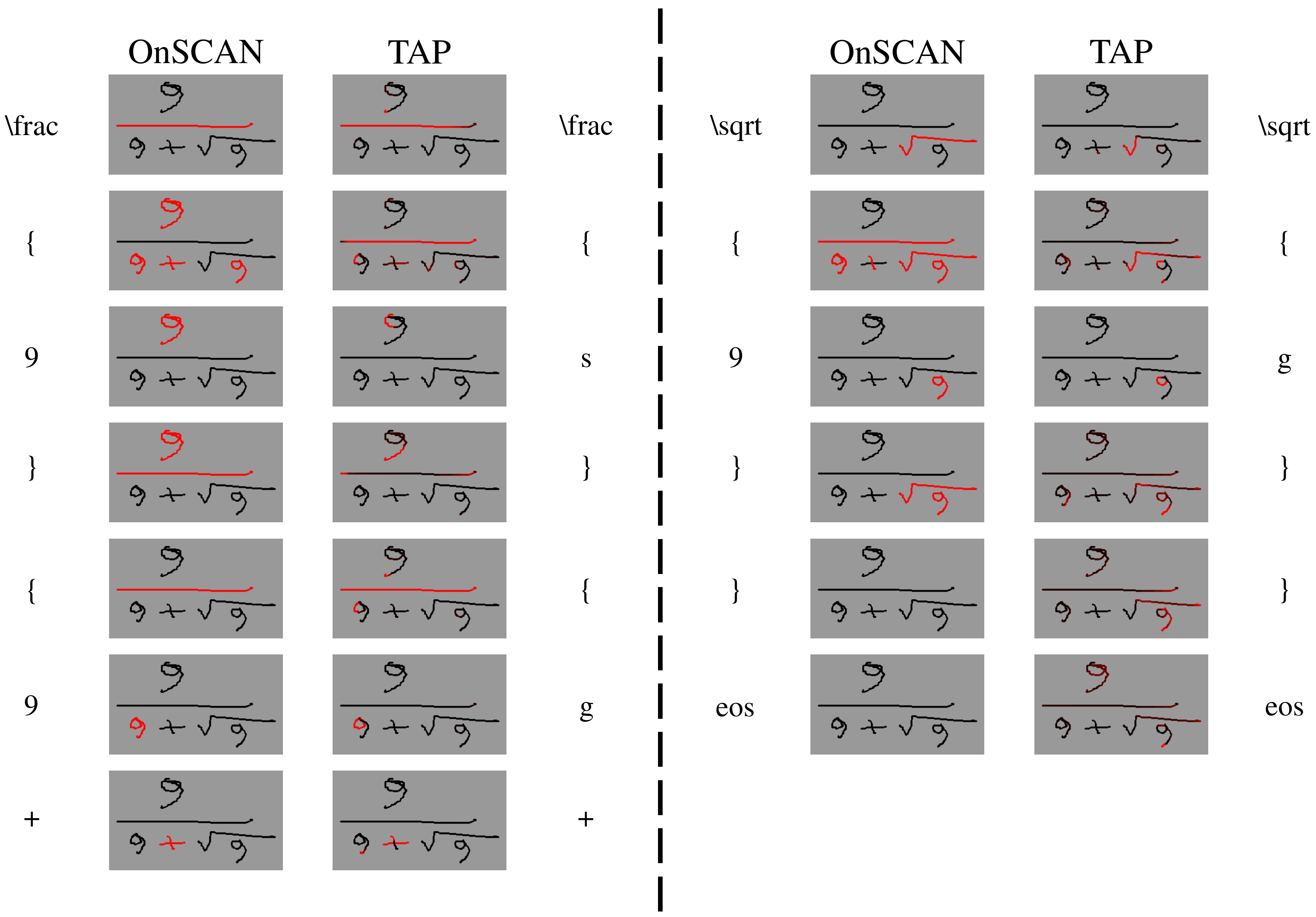}}
\caption{The attention visualization and recognition result comparison between OnSCAN and TAP for one handwritten mathematical expression with the LaTeX ground truth `` $\backslash$frac \{ 9 \} \{ 9 + $\backslash$sqrt \{ 9 \} \} ''.}
\label{online-attention}
\end{figure}
In Section~\ref{sec:singleModalityExperiment} and Section~\ref{sec:multiModalityExperiment}, we have demonstrated that SCAN can improve the performance of online HMER in both single-modality and multi-modality. In this section, we further show that SCAN can acquire more accurate symbol segmentation which is performed by attention. Moreover, the advantage of encoder fusion over the decoder fusion is explained by attention visualization.

We first compare the attention and recognition results of OnSCAN and TAP of one handwritten mathematical expression with the LaTeX ground truth `` $\backslash$frac \{ 9 \} \{ 9 + $\backslash$sqrt \{ 9 \} \} '' in Figure~\ref{online-attention}. It is obvious that OnSCAN correctly recognizes the example expression while TAP fails. Specifically, OnSCAN can focus on the exact points of the current predicted symbol at each time step, which in fact achieves accurate symbol segmentation and thus generates correct recognition result. As for TAP, it can only focus on parts of the points belonging to the current symbol. Besides, it will improperly focus on some redundant points belonging to other symbols. Therefore, TAP mistakenly recognizes the first ``9'' as ``s'' and the second/third ``9'' as ``g''. It is reasonable as the attended parts can be regraded as a part of ``9'', ``s'' or ``g''.

\begin{figure}[htb]
\centerline{\includegraphics[width=0.95\linewidth]{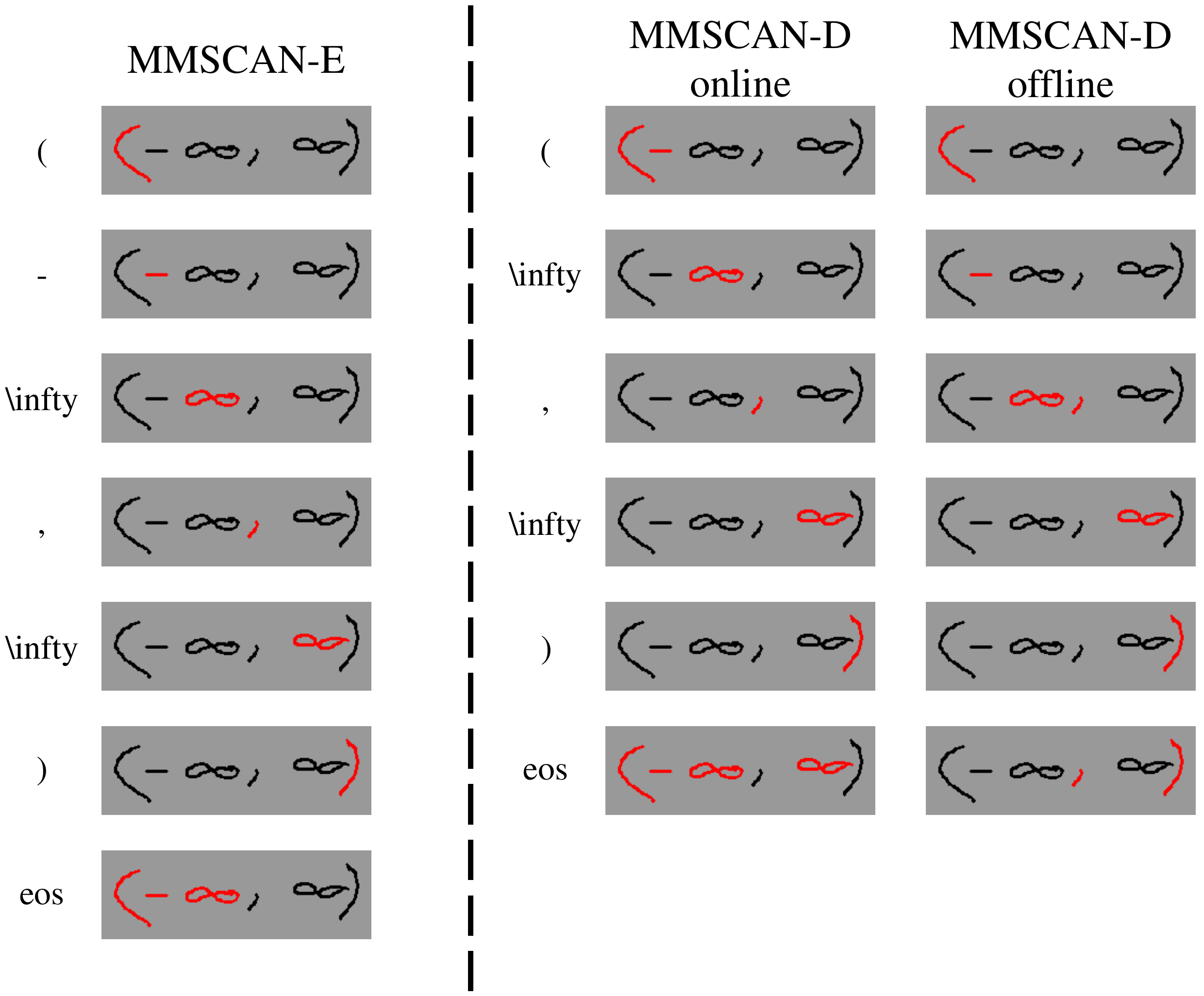}}
\caption{Attention and recognition results of MMSCAN-E and MMSCAN-D for one handwritten mathematical expression with the LaTeX ground truth `` ( - $\backslash$infty , $\backslash$infty ) ''. }
\label{mul-attention}
\end{figure}

The attention and recognition results of MMSCAN-E and MMSCAN-D for one handwritten mathematical expression are shown in Figure~\ref{mul-attention}. As the decoder of MMSCAN-D accepts both online and offline stroke-level features, accordingly attention results for both online and offline modalities are given. Ideally, the attention results of MMSCAN-D online and MMSCAN-D offline should be the same, namely the same attended strokes belonging to the symbol at each decoding step. However, as MMSCAN-D generates attention results over online and offline stroke-level features separately in the decoder, the attention results for online and offline modalities might be different leading to incorrect recognition. For example, at the first three steps of MMSCAN-D, the different attention results of online and offline modalities lead to a deletion error, namely incorrectly recognizing `` ( - $\backslash$infty '' as `` ( $\backslash$infty '' with the symbol ``-'' missing. On the contrary, MMSCAN-E generates exactly accurate attention results at each step and correctly recognizes the example expression. This indicates the superiority of early-stage fusion by better utilizing the alignments between online and offline modalities.

\begin{figure}[htb]
\centerline{\includegraphics[width=0.95\linewidth]{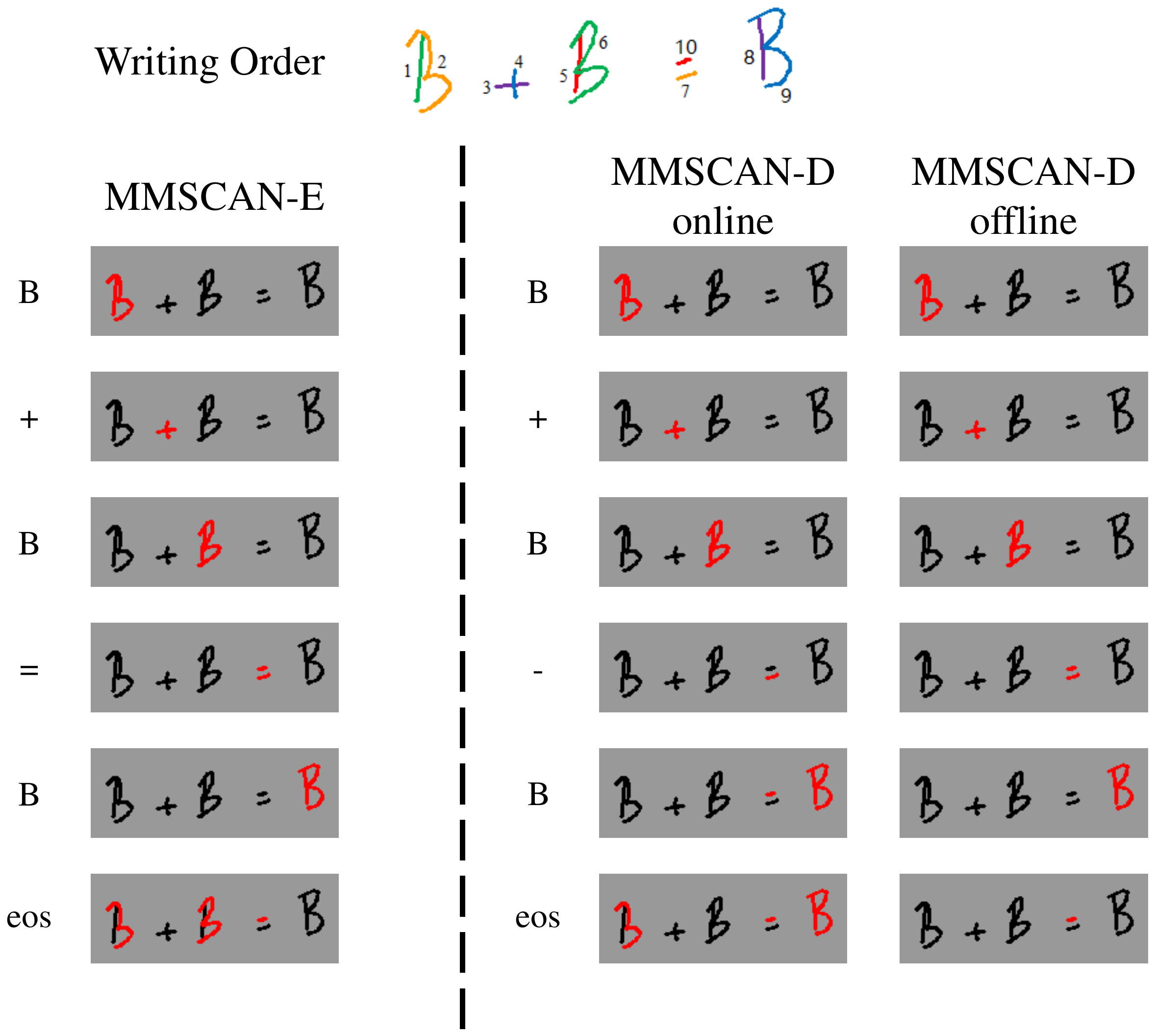}}
\caption{Attention visualization and recognition results of MMSCAN-E and MMSCAN-D for one handwritten mathematical expression with the LaTeX ground truth `` B + B = B ''. The problem of delayed strokes exists in this example.}
\label{mul-attention-order}
\end{figure}

One main motivation of multi-modal fusion for online HMER is to overcome problems in single modality by using information from both online and offline modalities. For example, a very common problem is caused by delayed strokes. As shown in Figure~\ref{mul-attention-order}, we take one expression with the LaTeX ground truth `` B + B = B '' as an example. The difficulty of recognizing this sample is that the writing order of this expression is different from the normal writing order. This expression consists of ten strokes with the corresponding writing order in Figure~\ref{mul-attention-order}. In general, after writing the second ``B'', we will write the ``='' by two strokes. However, in this example, one stroke (marked red) of the symbol ``='' is delayed to be written as the final stroke, which makes online modality difficult to correctly recognize. Although MMSCAN-D has information of two modalities and adopts re-attention to implement information interaction, it only has single modality information in pre-attention and still meets problems as the errors caused by pre-attention will be inherited in fine-attention. Therefore, MMSCAN-D incorrectly recognizes this expression as `` B + B - B''. Nevertheless, MMSCAN-E fuses two modalities in encoder, which authentically has both online and offline information when performing attention in the decoder. As a result, the global information in offline modality can help solve the delayed stroke problem and MMSCAN-E correctly recognizes this expression.

\subsection{Comparison of Recognition Speed (Q4)}
We compare the computational costs of whether employing SCAN in single-modal and multi-modal cases by investigating the test speed in this section. We present the total time cost (normalized by the time cost of TAP system) for recognizing the CROHME 2014 testing set in Table~\ref{tab:TestSpeed}. For the single modality, it is obvious that converting point-level/pixel-level features (TAP/WAP) into online/offline stroke-level features (OnSCAN/OffSCAN) can accelerate the testing procedure as the number of strokes is much smaller than the number of points/pixels, which reduces the computation cost of the decoder part. Similarly, in the multi-modal fusion case, MMSCAN-D is faster than E-MAN as MMSCAN-D replaces both point-level and pixel-level features with online and offline stroke-level features at the same time. Besides, MMSCAN-E with the highest ExpRate and StruRate can also achieve the best efficiency compared with MMSCAN-D and E-MAN due to the early-stage fusion. Overall, MMSCAN-E system only uses a half of time cost of E-MAN system and is even faster than offline WAP system.

\begin{table}[htb]
\caption{\label{tab:TestSpeed}{Comparison of time efficiency between SCAN and local (point-level or pixel-level) feature based approaches in both single-modal and multi-modal cases.}}
\centering
\begin{tabular}{l l c c c }
\toprule
Modality & System & ExpRate & StruRate & Time Cost\\
\midrule
\multirow{2}{*}{Online} &TAP & 48.47\% & 67.24\% & 1 \\
& OnSCAN & 51.22\% & 70.49\% & 0.91 \\
\midrule
\multirow{2}{*}{Offline} & WAP & 48.38\% & 70.08\% & 1.51\\
& OffSCAN & 47.67\% & 68.56\% & 1.28\\
\midrule
\multirow{3}{*}{Multi-modal}  & E-MAN & 54.05\% & 72.11\% & 2.66\\
& MMSCAN-D & 55.38\% & 71.30\% & 1.94\\
& MMSCAN-E & 57.20\% & 73.94\% & 1.32\\
\bottomrule
\end{tabular}
\end{table}

\section{Conclusion and Future Work}
In this study, we introduce a novel stroke constrained attention network (SCAN) for online handwritten mathematical expression recognition. The proposed model can be applied in both single-modal and multi-modal cases. We demonstrate through experimental results that SCAN can significantly improve the recognition performance and accelerate the testing procedure. Moreover, we verify that SCAN greatly improves the alignment via the attention visualization. In the future, we aim to investigate a better approach for fusing features from different modalities to acquire a more reasonable representation. Furthermore, we will perform the explicit symbol segmentation by using attention results.


\section*{References}

\bibliography{reference}

%








\end{document}